\documentclass[11pt,a4paper]{article}
\usepackage[hyperref]{acl2020}
\usepackage{times}
\usepackage{latexsym}

\usepackage{adjustbox}
\usepackage{multirow}
\usepackage{float}

\usepackage{microtype}

\aclfinalcopy 

\title{Growing Together: Modeling Human Language Learning With \textit{n}-Best Multi-Checkpoint Machine Translation}
\author{El Moatez Billah Nagoudi$^1$, Muhammad Abdul-Mageed$^1$,  Hasan Cavusoglu $^2$ \\
$^1$ Natural Language Processing Lab\\ 
    $^2$ Sauder School of Business\\
      $^{1,2}$ The University of British Columbia  \\
     \small{$^1$ \{muhammad.mageed,moatez.nagoudi\}@ubc.ca, $^2$ cavusoglu@sauder.ubc.ca}}

\date{}

\begin{document}
\maketitle
\begin{abstract}
We describe our submission to the 2020 Duolingo Shared Task on Simultaneous Translation And Paraphrase for Language Education (STAPLE)~\cite{staple20}. We view MT models at various training stages (i.e., checkpoints) as human learners at different levels. Hence, we employ an ensemble of multi-checkpoints from the same model to generate translation sequences with various levels of fluency. From each checkpoint, for our best model, we sample \textit{n}-Best sequences ($n=10$) with a beam width $=100$. We achieve $37.57~macro\ F_1$ with a $6$ checkpoint model ensemble on the official English to Portuguese  shared task test data, outperforming a baseline Amazon translation system of $21.30~ macro\ F_1$ and ultimately demonstrating the utility of our intuitive method.  

\end{abstract}

\section{Introduction}

\begin{figure}[t]
\begin{centering}
 \frame{\includegraphics[width=7.5cm, height=4.5cm]{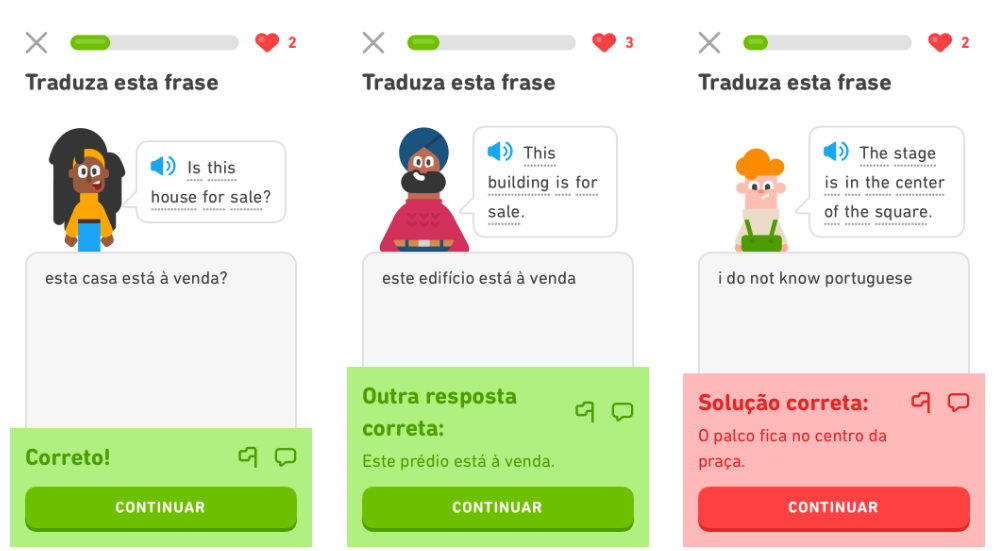}}
  \caption{ Translations proposed by English language learners at various levels of fluency, from diverse backgrounds. Our multi-checkpoint ensemble models mimic learner fluency.\textcolor{blue}{$^4$}
  }
  \label{dulingo}
\end{centering}
\end{figure}

Machine Translation (MT) systems are usually trained to output a single translation. However, many possible translations of a given input text can be acceptable.  This situation is common in online language learning applications such as \textit{Duolingo},\footnote{\url{https://www.duolingo.com/}} \textit{Babbel}\footnote{\url{https://www.babbel.com/}}, and \textit{Busuu}.\footnote{\url{https://www.busuu.com/}}  In applications of this type, learning happens via translation-based activities while evaluation is performed by comparing learners' responses to a large set of human acceptable translations. Figure~\ref{dulingo} shows an example of a typical situation extracted from the Duolingo application. 

The main set up of the 2020 Duolingo Shared Task on Simultaneous Translation And Paraphrase for Language Education (STAPLE 2020)~\cite{staple20} is such that one starts with a set of English sentences (prompts) and then generates high-coverage sets of plausible translations in the five target languages: Portuguese, Hungarian, Japanese, Korean,  and Vietnamese. For instance, if we want to translate the English (En) sentence \textit{``is my explanation clear?"} to Portuguese (Pt), all the translated Portuguese sentences illustrated in Table~\ref{pt_tran} would be acceptable.\footnote{Examples taken from shared task description at: \url{https://sharedtask.duolingo.com/}.} 

\noindent\textbf{Limited training data.} One challenge for training a sufficiently effective model we faced is the limited size of the source training data released by organizers ($4,000$ source English sentences coupled with $226,466$ Portuguese target sentences). We circumvent this limitation by training a model on a large dataset acquired from the OPUS corpus (as described in Section~\ref{sec:data}), which gives us a powerful MT system that we build on (see Section~\ref{subs:Basic}). We then exploit the STAPLE-provided training data in multiple ways (see Sections~\ref{subs:extended} and~\ref{subs:fine-tuned}) to extend this primary model as a way to nuance the model to the shared task domain. 

\noindent\textbf{Paraphrase via MT.} In essence, the shared task is a mixture of MT and paraphrase. This poses a second challenge: there is no paraphrase dataset to train the system on. For this reason, we resort to using outputs from the MT system in place of paraphrases. This required generating multiple sentences for each source sentence. To meet this need, we generate multiple translation hypotheses ($n$-Best) using a wide beam search (Section~\ref{subs:n-best}), perform `round-trip' translations exploiting these multiple outputs (Section~\ref{subs:paraphrasing}), and employ ensembles of checkpoints (Section~\ref{subs:multi-checkpoint}). 

\noindent\textbf{Diverse outputs.} A third challenge is that the target Portuguese sentences provided for training by organizers are produced by learners of English at various levels of fluency. This makes some of these Portuguese translations inarticulate (i.e., not quite fluent). MT systems are not usually trained to produce inarticulate translations (part of the time), and hence we needed to offer a solution that matches the different levels of language learners who produced the translations. Intuitively, we view MT systems trained at various stages (i.e., checkpoint) as learners with various levels of fluency. As such, we employ an ensemble of checkpoints to generate translations matching the different levels of learner fluency (Section~\ref{subs:multi-checkpoint}). Ultimately, \textbf{\textit{our contributions lie in alleviating the 3 challenges listed above}}.

The remainder of the paper is organized as follows: Section~\ref{sec:lit} is a brief overview of related work. In Section~\ref{sec:data}, we describe the data we use for both training and fine-tuning our models. Section~\ref{sec:Sys} presents the proposed MT system. Section~\ref{sec:app} describes our different methods. We discuss our results in Section~\ref{sec:test}, and conclude in Section~\ref{sec:con}.  \\

\begin{table}[h]
    \centering 
   
        \begin{tabular}{l|l}
        \hline    \hline
        \textbf{\small{English sentence}} &  is my explanation clear? \\ \hline
        \hline
             & - minha explicação está clara?   \\
             \textbf{\small{Accepted} }   & - minha explicação é clara?  \\ 
              \textbf{\small{Portuguese} }   &  - a  minha explicação é clara?  \\ 
            \textbf{\small{Translations}  } &  - está clara minha explicação?  \\ 
             & - minha explanação está clara? \\
            & - é clara minha explicação? \\
             \hline  
               \hline    
        \textbf{\small{English sentence}} &  you look so pretty! \\ \hline
        \hline
             & - você está tão linda!  \\
             \textbf{\small{Accepted} }   & - você está tão bonita! \\ 
              \textbf{\small{Portuguese} }   &  - você está muito linda! \\ 
            \textbf{\small{Translations}  } &  - você está muito bonita!  \\ 
             & - você parece tão linda!  \\
            & -  você parece tão bonita!\\
             \hline  
        \end{tabular}
    \caption{English sentences with their Portuguese translation samples from shared task training split. }
    \label{pt_tran}
\end{table}

\section{Related Work}\label{sec:lit}
We focus our related work overview on the task of paraphrase generation and its intersection with machine translation. Paraphrasing is the task of expressing the same textual units (e.g. sentence) with alternative forms using different words while keeping the original meaning intact. \footnote{\url{https://dictionary.cambridge.org/dictionary/english/paraphrase}} Over the last few years, MT has been the dominant approach for paraphrase generation. For instance, ~\citet{barzilay2001extracting,pang2003syntax} use multiple translations of the same text to train a paraphrase system. Similarly, \citet{bannard2005paraphrasing} use an MT phrase table to mapping an English sentences to various non-English sentences.

More recently, advances in neural machine translation (NMT) have spurred interest in paraphrase generation~\cite{sutskever2014sequence, luong2015stanford,aharoni2019massively}. For example, \citet{prakash2016neural} employ a stacked residual LSTM network  to learn a sequence-to-sequence model on paraphrase data. A parpahrase model with  adversarial training is presented by~\cite{li2017paraphrase}.~\citet{wieting2017paranmt,iyyer2018adversarial} propose a translation-based paraphrasing system, which is based on NTM to translate one side of a parallel corpus. Paraphrase generation with pivot NMT is used by~\cite{mallinson2017paraphrasing,yu2018qanet}.

\begin{table*}[h]
\centering
\begin{adjustbox}{width=\textwidth}
\renewcommand{\arraystretch}{1.5}
{
\begin{tabular}{l|c|c|c|c|c}
\hline 
\textbf{Corpus} & \textbf{Content} & \textbf{Documents} & \textbf{Sentences} & \textbf{En. Words} & \textbf{Pt. Words} \\ \hline \hline

\textbf{ParaCrawl v5}  & Parallel corpora from Web Crawls collected in the ParaCrawl project & 287 & 13.9M & 341.4M & 347.9M \\ 

\textbf{TildeMODEL v2018}  & This is the Tilde MODEL Corpus – Multilingual Open Data for European Languages & 6 & 3.6M & 134.1M & 100.4M \\ 

\textbf{DGT }  & A collection of translation memories provided by the JRC & 287 & 13.9M & 341.4M & 347.9M \\ 

\textbf{SciELO}  & Parallel corpus of full-text articles in Portuguese, English and Spanish from SciELO & 2 & 3.1M & 92.8M & 95.4M \\ 

\textbf{OpenSubtitles}  & A new collection of translated movie subtitles & 42,755 & 35.5M & 283.4M & 248.9M \\ 

\textbf{Tanzil} & A collection of Quran translations & 15 & 0.1M & 2.8M & 2.4M \\ 
\textbf{News Commentary} & A parallel corpus of News Commentaries provided by WMT & 7,185 & 0.6M & 15.4M & 15.5M  \\ 
\textbf{ Europarl v8} &  A parallel corpus extracted from the European Parliament web site & 10,344 & 2.0M & 59.5M & 6.1M \\ 

\textbf{JW300 v1 }  & JW300 is a parallel corpus of over 300 languages & 26,991 & 2.2M & 40.0M & 40.8M \\ 

\textbf{CAPES v1  }  & Parallel corpus of theses and dissertation abstracts in Portuguese and English from CAPES & 1 & 1.2M& 38.4M & 39.1M \\ 

\textbf{EMEA v3 }  & A parallel corpus  from the European Medicines Agency & 1,921 & 1.1M & 12.0M & 16.4M \\ 

\textbf{QED v2.0a }  & Open multilingual collection of subtitles for educational videos and lectures & 4,618 & 0.5M & 8.7M & 7.4M \\ 

\textbf{ JRC-Acquis v3.0} &  JRC-Acquis is a collection of legislative text of the European Union  & 20,507 & 1.7M & 64.3M & 64.8M \\ 

\textbf{Wikipedia} & A corpus of parallel sentences from Wikipedia & 5 & 1.8M & 47.0M & 44.8M   \\ 
\textbf{TED2013} & A parallel corpus of TED talk subtitles by CASMACAT & 1& 0.2M& 3.1M &2.9M    \\ 
\textbf{GNOME.} & A parallel corpus of GNOME localization files & 1,307 & 0.6M & 2.6M & 3.7M\\ 
\textbf{Tatoeba} & A collection of translated sentences from Tatoeba & 1 & 0.2M & 11.0M & 2.7M \\ 

\textbf{ECB v1} &Website and documentatuion from the European Central Bank & 1 & 0.2M & 5.8M & 6.2M \\ 

\textbf{bible-uedin v1} & Multilingual parallel corpus created from translations of the Bible & 2 & 62.2K & 1.8M & 1.7M \\ 

\textbf{GlobalVoices} &  A parallel corpus of news from the Global Voices website & 5,133 &71.5k &2.3M &2.3M \\ 
\textbf{KDE4} &A parallel corpus of KDE4 system messages & 2,136 & 0.2M & 2.4M &2.7M \\ 
\textbf{Ubuntu} & A parallel corpus of the Ubuntu Dialogue Corpus& 449 & 0.1M & 0.7M & 0.5M \\

\textbf{EUconst v1} & A parallel corpus collected from the European Constitution & 47 & 10.9K & 0.2M & 0.2M \\ 

\textbf{Books v1} & A collection of copyright free book& 1 & 1.4K & 33.8K & 32.3K \\ \hline

\textbf{Total} & All corpora extracted from OPUS & 162,425 & 77.7M & 1.5B & 1.4B \\ \hline 
\end{tabular}}
\end{adjustbox}
\caption{English-Portuguese datasets from \newcite{OPUS} used in our training.}
\label{tab1}
\end{table*}

\section{Data}\label{sec:data}

\subsection{Shared task data} 
As part of the STAPLE 2020 shared task, only \texttt{training} data were released. The target training split is a total of $526,466$ of learner translations of $4,000$ input (source) English sentences. We note that the number of translations of each English sentence varies, with an average of $\sim 132$ Portuguese target sentences for each English source sentence. As shared task organizers point out, this training dataset can be used as a reference/anchor points, and also serves as a strong baseline. For \texttt{evaluation}, a sets of $60,294$ translations (learner-crafted sentences) of 500 input English sentences were available on Colab. \texttt{Test} data were also made available only via Colab and comprised 500 English sentences learner-translated into $67,865$ Portuguese sentences. For all training, development, and test data, these translations are ranked and weighted according to actual learner response frequency. We refer the reader to the shared task description for more information.~\footnote{ \url{https://sharedtask.duolingo.com/\#data}.}

\subsection{OPUS data} 
In order to develop efficient  English-Portuguese MT models that can possibly work across different text domains, we make use of a large dataset of parallel English-Portuguese sentences extracted from the Open Parallel Corpus Project (OPUS)~\cite{OPUS}. OPUS\footnote{\url{http://opus.nlpl.eu/}} contains more than $2.7$ billion parallel sentences in $90$ languages. The specific corpus we extracted consists of data from multiple domains and sources including: ParaCrawl project~\cite{espla2019paracrawl}, EUbookshop~\cite{skadicnvs2014billions}, Tilde Model~\cite{rozis2017tilde}, translation memories (DGT) \cite{steinberger2013dgt}, Open-Subtitles~\cite{creutz2018open}, SciELO Parallel  \cite{soares2018large}, JRC-Acquis Multilingual~\cite{steinberger2006jrc},  Tanzil~\cite{zarrabizadeh2007tanzil}, Europarl Parallel~\cite{koehn2005europarl}, TED 2013~\cite{cettoloEtAl:EAMT2012}, Wikipedia~\cite{wolk2014building}, Tatoeba~\footnote{\url{www.tatoeba.org}}, QCRI Educational Domain~\cite{abdelali2014amara}, GNOME localization files,~\footnote{\url{www.10n.gnome.org}} Global Voices,~\footnote{\url{www.globalvoices.org/}} KDE4,~\footnote{\url{www.i18n.kde.org}}, Ubuntu,~\footnote{\url{www.translations.launchpad.net}} and Multilingual  Bible~\cite{christodouloupoulos2015massively}.  To train our models, we extract more than $77.7M$ parallel (i.e., English-Portuguese) sentences from the whole collection. The extracted dataset comprises more than $1.5B$  English tokens and $1.4B$  Portuguese tokens. More details about the training dataset are given in Table~\ref{tab1}.

\subsection{Pre-Processing} Pre-processing is an important step in building any MT model as it can significantly affect the end results. We remove punctuation and tokenize all data with the Moses tokenizer \cite{koehn2007open}. We also use joint Byte-Pair Encoding (BPE) with 60K split operations for subword segmentation~\cite{sennrich2015neural}. 

\section{Models}
\label{sec:Sys}
In this section, we first describe the architecture of our models. We then explain the different ways we train the models on various subsets of the data.

\subsection{Architecture}
 Our models are  mainly based on a Convolutional Neural Network (CNN) architecture~\cite{kim2014convolutional, gehring2017convolutional}. This convolutional architecture  exploits  BPE~\cite{sennrich2015neural}. The architecture is as follows: 20 layers in the encoder and 20 layers in the decoder, a multiplicative attention~\cite{luong2015effective} in every decoder layer, a kernel  width of 3  for both the encoder and the decoder, a hidden size 512, and an embedding size of 512, and 256 for the encoder and decoder layers respectively. We use a Fairseq implementation~\cite{ott2019Fairseq}.


\subsection{Basic En$\leftrightarrow$Pt Models} 
\label{subs:Basic}
We trained  two MT models, English-to-Portuguese (En$\rightarrow$Pt) and Portuguese-to-English (Pt$\rightarrow$En), on 4 V100 GPUs, following the setup described in~\citet{ott2018scaling}. For both models, the learning rate was set to $0.25$, a dropout of $0.2$, and a maximum tokens of $4,000$ for each mini-batch.  We train our models on  the $77.7M$ parallel sentences of the OPUS dataset described in Section~\ref{sec:data}. Validation is performed on the development data from STAPLE 2020~\cite{staple20}.

\subsection{En$\rightarrow$Pt Extended Model} 

\label{subs:extended} We use the training data of the STAPLE 2020 shared task\footnote{\url{http://sharedtask.duolingo.com/\#data}} to create a new En-Pt parallel  dataset. More specifically, at the \textit{target} side, we use all the Portuguese gold translations while duplicating the same English source sentence at the \textit{source} side. This results in a new training set of $251,442$ En-Pt parallel sentences. We refer to this training dataset as STAPLE-TRAIN, or simply \textit{S-TRAIN}. We then merge OPUS and S-TRAIN to train an En$\rightarrow$Pt model from scratch. We refer to this new model as the \textit{extended model}. 

\subsection{En$\rightarrow$Pt Fine-Tuned Model} \label{subs:fine-tuned}
Fine-tuning with domain-specific data, from a domain of interest, can be an effective strategy when it is desirable to develop systems for such a domain~\cite{ott2019Fairseq,ott2018scaling}. Motivated by this, we experiment with using the STAPLE-based S-TRAIN parallel dataset from the previous sub-section to fine-tune our En$\rightarrow$Pt \textit{basic} model for $5$ epochs.~\footnote{We choose the number of epochs arbitrarily, but note that it is a hyper-parameter that can be tuned.} We will refer to the model resulting from this fine-tuning process simply as the \textit{fine-tuned model}.

\begin{figure*}[t]
\begin{centering}
 \includegraphics[scale=0.65]{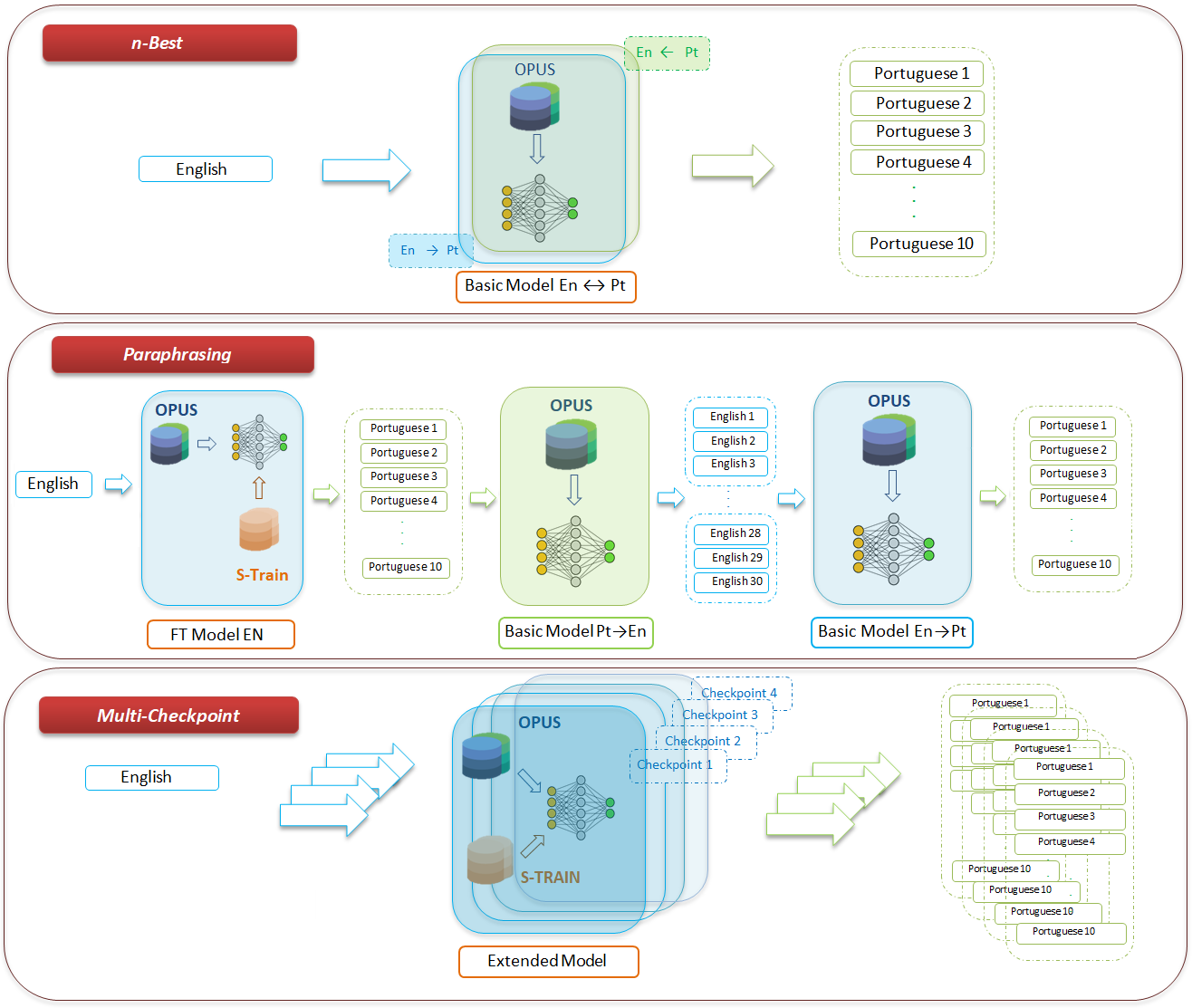} 
  \caption{An illustration of our proposed models and methods: \textbf{(a) \textit{n}-Best prediction method} with $n=10$ resulting in the En$\rightarrow$Pt \textit{basic} model; \textbf{(b) paraphrasing method} with $n=10$ and $n'=3$ used in the En$\rightarrow$Pt \textit{fine-tuning} and the En$\leftrightarrow$Pt basic models, \textbf{(c) multi-checkpoint method} used with $n=10$ and $m=4$ for the En$\rightarrow$Pt \textit{extended} model.}
\end{centering}
\end{figure*} 

\section{Model Deployment Methods} 
\label{sec:app}
In order to enhance the $1$-to-$n$ En-Pt translation, we propose three methods based on the previously discussed MT models (see section~\ref{sec:Sys}). These methods are \textit{$n$-Best prediction}, \textit{multi-checkpoint translation}, and \textit{paraphrasing}.

\subsection{ \textit{n}-Best Prediction } 
\label{subs:n-best}
We first use our three MT models (\textit{basic}, \textit{extended}, and \textit{fine-tuned}) with a beam search size of $100$ to generate  $n$-Best translation hypotheses. We then use the average log-likelihood to score each of these hypotheses. Finally, we select the hypothesis with the $n$ highest score as our output. 
 
\subsection{ Paraphrasing  } 
\label{subs:paraphrasing}
Paraphrasing is an effective data augmentation method which is commonly used in MT tasks \cite{poliak2018evaluation, iyyer2018adversarial}. In order to extend the list of accepted Portuguese translations, we use both of our En$\rightarrow$Pt and  Pt$\rightarrow$En models, as follows: 

\begin{enumerate}
    \item Translate the English sentences using the  En$\rightarrow$Pt model. For instance, we generate $n$-Best ($n=10$) Portuguese sentences for each English source sentence.
    \item Then, we use the Pt$\rightarrow$En model to get  $n'$-Best English translations (we experiment with $n'=1$, $3$, and $5$) for each of the $10$ Portuguese sentence. At this point, we would have $10*n'$ new English sentences (oftentimes with duplicate generations that we remove). These new sentences represent paraphrases of the original English sentence. \\ 
    \item After de-duplication, the  new  English sentences are fed to the En$\rightarrow$Pt model to get the $1$-Best Portuguese translation. 
\end{enumerate}

\subsection{Multi-Checkpoint Translation} 
\label{subs:multi-checkpoint}
Our third method is based on saving the models at given epochs (checkpoints) during training. We use the $m$ last checkpoints (models) to generate the $n$-Best translation hypotheses (the same way as our $n$-Best prediction method). We then de-duplicate the outputs of all the $m$ models and use them in evaluation. We now describe our evaluation.   

\section{Evaluation}
\label{sec:test}

In order to evaluate our methods, we carry out a number of experiments. First, we consider performance of each proposed method on the official \texttt{training} and \texttt{development} datasets of STAPLE~\cite{staple20}. Our models were ultimately evaluated on the shared task \texttt{test} data. We now describe STAPLE evaluation metrics and baselines as provided by organizers, before reporting on our results on training, development, and test.

\subsection{Evaluation Metrics \& Baselines} 

\vspace{0.5em}
\textbf{Weights of Translation.} We note that each Portuguese translated sentence has a weight as provided in the gold dataset. The weights of translations correspond to user (learner) response rates. These weights are used primarily for scoring. The STAPLE 2020 shared task data takes the format illustrated in Table~\ref{pt_tran_wieg}.  

\begin{table}[h]
    \centering 

        \begin{tabular}{l|l}
        \hline    \hline
        \multicolumn{2}{l}{\textbf{{English Sentence}} :\small{~~~~is my explanation clear?}} \\ \hline
 
           \bf \small Weights & \small{\textbf{Portuguese Translation}} \\ \hline
             $0.26739$ & - minha explicação está clara?   \\
              $0.16168$ & - minha explicação é clara?  \\ 
             $0.11109$& - a  minha explicação é clara? \\ 
              $0.08778$ & - está clara minha explicação?  \\ 
            $0.05717$  &  - minha explanação está clara? \\
             \hline  
               \hline        \multicolumn{2}{l}{\textbf{{English Sentence}} :\small{~~~~this is my fault.}}\\  \hline
 
           \bf \small Weights & \textbf{\small{Portuguese translation}} \\ \hline
             $0.17991$ & - isto é minha culpa.   \\
              $0.10664$ & - isso é minha culpa. \\ 
             $0.08944$& - esta é minha culpa. \\ 
              $0.07794$ & - isto é culpa minha.  \\ 
            $0.06803$  &  - é minha culpa. \\
             \hline  
               \hline 
        \end{tabular}
    \caption{English sentences with their Portuguese translation and  Weights samples from shared task train data. }
    \label{pt_tran_wieg}
\end{table}

\noindent\textbf{Metrics.} 
Performance of MT systems in the shared task is quantified and scored based on how well a model can return all human-curated acceptable translations, weighted by the likelihood that an English learner would respond with each translation \cite{staple20}. As such, the main scoring metric is the \textit{weighted macro} $F_{1}$, with respect to the accepted translations. To compute \textit{weighted macro} $F_{1}$  (see formula~\ref{Macro_F1}), the weighted $F_{1}$ for each English sentence (s) is calculated and the average over all the sentences in the corpus is computed. The weighted $F_{1}$ (see formula~\ref{W_F1}) is computed using the unweighted precision (see formula~\ref{Precision}) and the weighted recall (see formulas~\ref{WTP},~\ref{WFN} and \ref{w_Recall}). \\

\small
\begin{equation}
Precision\ (s) = \frac{TP_{s}}{ TP_{s}+ FN_{s}}
\label{Precision}
\end{equation}

\small
\begin{equation}
WTP_{s}=  \sum_{s\in TP_{s}}{weight(t)}
\label{WTP}
\end{equation}
\normalsize
\small
\begin{equation}
WFN_{s}=  \sum_{s\in FN_{s}}{weight(t)}
\label{WFN}
\end{equation}
\normalsize

\small
\begin{equation}
Weighted\ Recall\ (s) = \frac{WTP_{s}}{ WTP_{s}+ WFN_{s}}
\label{w_Recall}
\end{equation}

\small
\begin{equation}
Weighted\ F1(s)   = \frac{2\cdot Prec.\ (s) \cdot W.\ Recall\ (s) }{ Prec.\ (s) + W.\ Recall\ (s)}
\label{W_F1}
\end{equation}

\small
\begin{equation}
Weighted\ Macro\ F_{1}   = \sum_{s\in S}{\frac{Weighted\ F1(s)}{|S|}} \\
\label{Macro_F1}
\end{equation}

\normalsize

\noindent\textbf{Baselines.} We adopt the two baselines offered by the task organizers. These are based on Amazon and Fairseq translation systems and are at $21.30\%$ and $13.57\%$, respectively. More information about these baselines can be reviewed at the shared task site listed earlier.

\subsection{Evaluation on TRAIN and DEV}

In this section, we report the results of our 3 proposed methods, \textit{ (a) n-Best prediction}, \textit{(b) paraphrasing}, and \textit{(c) multi-checkpoint translation} using the MT models presented in section~\ref{sec:Sys}.

\noindent\textbf{Evaluation on TRAIN.} For (a) the \textit{n}-Best prediction method, we explore the 4 different values of $n$ in the set \{5,\ 10,\ 15, 20\}. For (b) the paraphrase method, we set the number of Portuguese sentences to $n'=\{1,\ 3,\ 5\}$. Finally, (c) the multi-checkpoint method was tested with 4 different values for the number of checkpoints $m = \{2,\ 4,\ 6,\ 8\}$. \\ For paraphrasing and multi-checkpoint translation, we fix the number of \textit{n}-best translations $n$ to $10$, varying the values of $n'$ and $m$ only when evaluating our \textit{extended} model. This leads us to identifying the best evaluation values of $n'= 3$ and $m = 6$, which we then use when evaluating our \textit{basic} and \textit{fine-tuned} models. \\ 

\noindent\textbf{Evaluation on DEV.} For evaluation on the STAPLE development data, we adopt the same procedure followed for evaluation on the train split. Table~\ref{tab:res-dev} summarizes our experiments with different configurations (i.e., values of $n$, $n'$, and $m$ ) on train and development task data, respectively. \\

\begin{table*}[ht]
 \begin{center}
\begin{tabular}{lc|c|c|c||c|c|c||c|c|c}
\cline{3-11} 
                                     & \multicolumn{10}{c}{\bf Train Data}                                                 \\
                                      \cline{3-11} 
          &   \multicolumn{4}{c||}{ \bf \small  Basic Model } & \multicolumn{3}{c||}{ \bf  \small  Extended Model } & \multicolumn{3}{c}{ \bf \small   Fine-Tuned Model} \\ \hline \hline
\multicolumn{1}{l|}{ \bf  \small  Method}  & \bf n &  \small  Prec.  &   \small   W. Recall  &  \small  W. $F_{1}$     &    \small   Prec.  &     \small  W. Recall  &  \small W. $F_{1}$   &   \small  Prec.  &    \small   W. Recall  &  \small  W. $F_{1}$    \\ \hline
\multicolumn{1}{l|}{\multirow{5}{*}{\bf  \small  \textit Prediction}} & 5 &   55.44 & 23.87 &27.41 & 45.51 &28.24 & 29.43& 44.68
& 26.91 & 28.38      \\ 
\multicolumn{1}{l|}{\small \textbf{~~~\textit{n}-Best}}                  & 10 &     42.78& 29.65 &28.47 &46.02&	34.18	&\bf33.51&41.81 & 32.19 & 30.33      \\ 
\multicolumn{1}{l|}{}                  & 15  &        37.42& 27.09 & 29.17 &39.25&	35.50&	31.80&45.51&28.24&29.43       \\ 
\multicolumn{1}{l|}{}                  &  20  &     29.68& 38.24&27.48&39.22&35.49&31.79&46.23&27.04&30.27       \\  \hline \hline

\multicolumn{1}{l|}{\multirow{5}{*}{\bf  \small  Paraphrasing}} & \bf $n^\prime$ & \small  Prec.  &   \small   W. Recall  &  \small  W. $F_{1}$     &    \small   Prec.  &     \small  W. Recall  &  \small W. $F_{1}$   &   \small  Prec.  &    \small   W. Recall  &  \small  W. $F_{1}$    \\ \cline{2-11} 
\multicolumn{1}{l|}{}                  & 1 &    -   &   -    &     -  &  40.24     & 35.01      &  31.68     &     -  &   -    &    -   \\ 
\multicolumn{1}{l|}{}                  & 3  &   40.24    &  35.01     &     31.68  &    46.45   & 35.08      &  \bf 34.39     & 39.98      &    37.27   &32.89       \\ 
\multicolumn{1}{l|}{}                  &  5  &  -     &    -   &  -     &   40.44    & 39.20      &    34.16   &   -    &    -   &    -   \\ 

\hline\hline
\multicolumn{1}{l|}{} & \bf m & \small  Prec.  &   \small   W. Recall  &  \small  W. $F_{1}$     &    \small   Prec.  &     \small  W. Recall  & \small W. $F_{1}$   &   \small  Prec.  &    \small   W. Recall  &  \small  W. $F_{1}$    \\ \cline{2-11} 
\multicolumn{1}{l|}{\multirow{5}{*}{\bf  \small Checkpoint}}                  & 2 &    -   &    -   &   -    &   58.81    &   31.57    &  36.57      &   -    &  -     &    -   \\ 
\multicolumn{1}{l|}{\small~~~~\textbf{ Multi}}                  & 4  &     -  &   -   &    -   &    50.53    &   44.22    &    40.76   &    -   &   -    &     -  \\ 
\multicolumn{1}{l|}{}                  &  6  &   44.44    & 45.52       & 39.46      &   49.58    &  44.92     &     \bf 40.78  &    36.77   &    52.73   &    38.54   \\ 
\multicolumn{1}{l|}{}                  &  8  &   -    &   -    &    -   &     42.16  &    44.96   &   39.28    &  -     &     -  &    -   \\ 
\hline\hline



                         & \multicolumn{10}{c}{\bf DEV Data}                                                 \\
                                       \cline{3-11} 

 &   \multicolumn{4}{c||}{ \bf \small  Basic Model } & \multicolumn{3}{c||}{ \bf  \small  Extended Model } & \multicolumn{3}{c}{ \bf \small   Fine-Tuned Model} \\ \hline \hline
\multicolumn{1}{l|}{ \bf  \small  Method}  & \bf n &  \small  Prec.  &   \small   W. Recall  &  \small  W. $F_{1}$     &    \small   Prec.  &     \small  W. Recall  &   \small W. $F_{1}$   &   \small  Prec.  &    \small   W. Recall  &  \small  W. $F_{1}$    \\ \hline
\multicolumn{1}{l|}{\multirow{5}{*}{\bf  \small Prediction }} & 5 &    -  &-&-&52.48&26.27&29.87 &  -     &  -     &    - \\ 
\multicolumn{1}{l|}{\small \textbf{~~~\textit{n}-Best}  }                  & 10 & 32.56&36.83&29.33&36.52&41.09&\bf32.96&35.39&37.84&31.30      \\ 
\multicolumn{1}{l|}{}                  & 15  &  -   &-&-&38.62&37.46&32.36&  -     &  -     &  -  \\ 
\multicolumn{1}{l|}{}                  &  20  &   -    &-&-&36.03&37.44&31.33&  -     &  -     &  -     \\  \hline \hline

\multicolumn{1}{l|}{\multirow{5}{*}{\bf  \small  Paraphrasing}} & \bf n' & \small  Prec.  &   \small   W. Recall  &  \small  W. $F_{1}$     &    \small   Prec.  &     \small  W. Recall  &   \small W. $F_{1}$   &   \small  Prec.  &    \small   W. Recall  &  \small  W. $F_{1}$    \\ \cline{2-11} 
\multicolumn{1}{l|}{}                  & 1 &  -     &  -     &  -     &  45.77     &    33.31   &    33.05   &  -     &  -     &   -    \\ 
\multicolumn{1}{l|}{}                  & 3  &   48.66    &31.17       &    32.43   &  46.34     &  33.85     &    33.17   &   39.98    &   37.27    &   32.89    \\ 
\multicolumn{1}{l|}{}                  &  5  &  -     &  -     &  & 46.14    &  34.26     &  \bf33.40     &  -     &  -     &      -      \\ 
\hline \hline

\multicolumn{1}{l|}{} & \bf m & \small  Prec.  &   \small   W. Recall  &  \small  W. $F_{1}$     &    \small   Prec.  &     \small  W. Recall  &  W. $F_{1}$   &   \small  Prec.  &    \small   W. Recall  &  \small  W. $F_{1}$    \\ \cline{2-11} 
\multicolumn{1}{l|}{\multirow{5}{*}{\bf  \small  Checkpoint}}                  & 2 &  -     &  -     &  -     &    55.88	& 32.16 &	35.26  &   -     &  -     &   -    \\ 
\multicolumn{1}{l|}{\small~~~~\textbf{ Multi}}                  & 4  &  -     &  -     &  -     & 52.27&37.35&38.25    &  -     &  -     &   -    \\ 
\multicolumn{1}{l|}{}                  &  6  &    45.35    &    43.20   &  38.04           &   56.42    &     37.31  &    39.16   &     45.01    &  41.23   &    37.26   \\ 
\multicolumn{1}{l|}{}                  &  8  &  -     &  -     &  -     &   53.83    &    38.85   &   \bf 39.21    &  -     &  -     &   -    \\ 
\hline\hline

\end{tabular}
  \end{center}
   \caption{\label{font-table}   Performance on the STAPLE 2020 Train and Dev data splits.}
    \label{tab:res-dev}
\end{table*}

\noindent \textbf{Discussion.}  
Results presented in Table~\ref{tab:res-dev} demonstrate that all the models with the different methods and configurations outperform the the official shared task baseline with $macro\ F_{1}$ scores between $27.41\%$ and $40.78\%$. As expected, fine-tuning the En$\rightarrow$Pt basic model with the S-TRAIN data-set improves the results with a mean of $+1.46\%$ on the training data. We also observe that training on the concatenated OPUS and S-TRAIN data-sets \textit{from scratch} leads to better results compared to the \textit{exclusive fine-tuning} method.

Based on these results, we can see that the best configuration is the multi-checkpoint method used with the \textit{extended} MT model. This configuration obtains the best  $macro\ F_{1}$ score of $40.78\%$ and $39.21\%$ on the training and development STAPLE data splits, respectively. 

\begin{table}[H]
 \begin{center}
\begin{tabular}{lc|c|c|c}
\cline{2-5} \cline{2-5} 
\cline{2-5} 
 &  \multicolumn{3}{c}{ \bf  \small  Extended Model } \\ \hline \hline
 
 \multicolumn{1}{l|}{ \bf  \small  Method}  & \bf m &  \small  Prec.  &   \small   W. Recall  &  \small  W. $F_{1}$ \\    \hline

\multicolumn{1}{l|}{\bf  \small  Aws Baseline}                  &  -  &   87.80  &   13.98  &   21.29       \\ \hline
\multicolumn{1}{l|}{\bf  \small  Fairseq Baseline}                  &  -  &  28.25  &   11.70  &   13.57       \\ \hline
\hline

\multicolumn{1}{l|}{\multirow{3}{*}{\bf  \small  Multi-Checkpoint}}               & 4  &     60.14   &   33.14    &  37.06      \\ 
\multicolumn{1}{l|}{}                  &  6    &   53.83   &     36.50  &    \bf37.57    \\ 
\multicolumn{1}{l|}{}                  &  8  &   49.94  &   38.27  &   37.21       \\ \hline   \hline

\end{tabular}
  \end{center}
   \caption{\label{font-table}   Results on STAPLE 2020 Test Data.}
    \label{tab:res-test}
\end{table}

\subsection{Evaluation on TEST}

In test phase, we submitted translations from 3 systems for the STAPLE English-Portuguese sub-task. The 3 systems are based on our \textit{multi-checkpoint translation} with the \textit{extended} model. The number of checkpoints used was $m = \{4,\ 6,\ 8\}$, and $n$  is fixed to $10$ (i.e., the best value of $n$ identified on training data with our \textit{extended} model). Table~\ref{tab:res-test} shows the results of our 3 final submitted systems as returned by the shared task organizers. \\

\noindent\textbf{Discussion.} Our results indicate that when the multi-checkpoint method with the extended model and only  two last checkpoints ($m=4$) is used, the $macro\ F_{1}$ score reaches $37.07\%$ (with a best precision of $60.14\%$). This method with $m=6$ represents our best $macro\ F_{1}$ score $37.57\%$ for the English-Portuguese translation sub-task. We note that with this configuration we outperform the Amazon and Fairseq translation baseline systems (at $+15.92\%$ and $+23.99\%$, respectively) provided by the task organizers. We also observe that when $m$ is set to $8$, the $macro\ F_{1}$ slightly decreases to $37.21\%$. Ultimately, our findings show the utility of using multiple checkpoint ensembles as a way to mimic the various levels of language learners. Simple as this approach is, we find it quite intuitive.


\section{Conclusion}
\label{sec:con}
In this work, we described our contribution to the 2020 Duolingo Shared Task on Simultaneous Translation And Paraphrase for Language Education (STAPLE)~\cite{staple20}. Our system targeted the English-Portuguese sub-task. Our models effectively make use of an approach based on $n$-Best prediction and multi-checkpoint translation. Our use of the OPUS dataset for training proved quite successful. In addition, based on our results, our intuitive deployment of a multi-checkpoint ensemble coupled with \textit{n}-Best decoded translations seem to mirror leaner proficiency. As future work, we plan to explore other methods on new language pairs.\\

\section*{Acknowledgements}
MAM gratefully acknowledge the support of the Natural Sciences and Engineering Research Council of Canada (NSERC), the Social Sciences Research Council of Canada (SSHRC), and Compute Canada (\url{www.computecanada.ca}).\\

\bibliography{acl2020}

\begin{thebibliography}{35}
\expandafter\ifx\csname natexlab\endcsname\relax\def\natexlab#1{#1}\fi

\bibitem[{Abdelali et~al.(2014)Abdelali, Guzman, Sajjad, and
  Vogel}]{abdelali2014amara}
Ahmed Abdelali, Francisco Guzman, Hassan Sajjad, and Stephan Vogel. 2014.
\newblock The amara corpus: Building parallel language resources for the
  educational domain.
\newblock In \emph{LREC}, volume~14, pages 1044--1054.

\bibitem[{Aharoni et~al.(2019)Aharoni, Johnson, and
  Firat}]{aharoni2019massively}
Roee Aharoni, Melvin Johnson, and Orhan Firat. 2019.
\newblock Massively multilingual neural machine translation.
\newblock \emph{arXiv preprint arXiv:1903.00089}.

\bibitem[{Bannard and Callison-Burch(2005)}]{bannard2005paraphrasing}
Colin Bannard and Chris Callison-Burch. 2005.
\newblock Paraphrasing with bilingual parallel corpora.
\newblock In \emph{Proceedings of the 43rd Annual Meeting on Association for
  Computational Linguistics}, pages 597--604. Association for Computational
  Linguistics.

\bibitem[{Barzilay and McKeown(2001)}]{barzilay2001extracting}
Regina Barzilay and Kathleen McKeown. 2001.
\newblock Extracting paraphrases from a parallel corpus.
\newblock In \emph{Proceedings of the 39th annual meeting of the Association
  for Computational Linguistics}, pages 50--57.

\bibitem[{Cettolo et~al.(2012)Cettolo, Girardi, and
  Federico}]{cettoloEtAl:EAMT2012}
Mauro Cettolo, Christian Girardi, and Marcello Federico. 2012.
\newblock Wit$^3$: Web inventory of transcribed and translated talks.
\newblock In \emph{Proceedings of the 16$^{th}$ Conference of the European
  Association for Machine Translation (EAMT)}, pages 261--268, Trento, Italy.

\bibitem[{Christodouloupoulos and
  Steedman(2015)}]{christodouloupoulos2015massively}
Christos Christodouloupoulos and Mark Steedman. 2015.
\newblock A massively parallel corpus: the bible in 100 languages.
\newblock \emph{Language resources and evaluation}, 49(2):375--395.

\bibitem[{Creutz(2018)}]{creutz2018open}
Mathias Creutz. 2018.
\newblock Open subtitles paraphrase corpus for six languages.
\newblock \emph{arXiv preprint arXiv:1809.06142}.

\bibitem[{Espl{\`a}-Gomis et~al.(2019)Espl{\`a}-Gomis, Forcada,
  Ram{\'\i}rez-S{\'a}nchez, and Hoang}]{espla2019paracrawl}
Miquel Espl{\`a}-Gomis, Mikel~L Forcada, Gema Ram{\'\i}rez-S{\'a}nchez, and
  Hieu Hoang. 2019.
\newblock Paracrawl: Web-scale parallel corpora for the languages of the eu.
\newblock In \emph{Proceedings of Machine Translation Summit XVII Volume 2:
  Translator, Project and User Tracks}, pages 118--119.

\bibitem[{Gehring et~al.(2017)Gehring, Auli, Grangier, Yarats, and
  Dauphin}]{gehring2017convolutional}
Jonas Gehring, Michael Auli, David Grangier, Denis Yarats, and Yann~N Dauphin.
  2017.
\newblock Convolutional sequence to sequence learning.
\newblock In \emph{Proceedings of the 34th International Conference on Machine
  Learning-Volume 70}, pages 1243--1252. JMLR. org.

\bibitem[{Iyyer et~al.(2018)Iyyer, Wieting, Gimpel, and
  Zettlemoyer}]{iyyer2018adversarial}
Mohit Iyyer, John Wieting, Kevin Gimpel, and Luke Zettlemoyer. 2018.
\newblock Adversarial example generation with syntactically controlled
  paraphrase networks.
\newblock \emph{arXiv preprint arXiv:1804.06059}.

\bibitem[{Kim(2014)}]{kim2014convolutional}
Yoon Kim. 2014.
\newblock Convolutional neural networks for sentence classification.
\newblock \emph{arXiv preprint arXiv:1408.5882}.

\bibitem[{Koehn(2005)}]{koehn2005europarl}
Philipp Koehn. 2005.
\newblock Europarl: A parallel corpus for statistical machine translation.
\newblock In \emph{MT summit}, volume~5, pages 79--86. Citeseer.

\bibitem[{Koehn et~al.(2007)Koehn, Federico, Shen, Bertoldi, Bojar,
  Callison-Burch, Cowan, Dyer, Hoang, Zens et~al.}]{koehn2007open}
Philipp Koehn, Marcello Federico, Wade Shen, Nicola Bertoldi, Ondrej Bojar,
  Chris Callison-Burch, Brooke Cowan, Chris Dyer, Hieu Hoang, Richard Zens,
  et~al. 2007.
\newblock Open source toolkit for statistical machine translation: Factored
  translation models and confusion network decoding.
\newblock In \emph{Final Report of the Johns Hopkins 2006 Summer Workshop}.

\bibitem[{Li et~al.(2017)Li, Jiang, Shang, and Li}]{li2017paraphrase}
Zichao Li, Xin Jiang, Lifeng Shang, and Hang Li. 2017.
\newblock Paraphrase generation with deep reinforcement learning.
\newblock \emph{arXiv preprint arXiv:1711.00279}.

\bibitem[{Luong and Manning(2015)}]{luong2015stanford}
Minh-Thang Luong and Christopher~D Manning. 2015.
\newblock Stanford neural machine translation systems for spoken language
  domains.
\newblock In \emph{Proceedings of the International Workshop on Spoken Language
  Translation}, pages 76--79.

\bibitem[{Luong et~al.(2015)Luong, Pham, and Manning}]{luong2015effective}
Minh-Thang Luong, Hieu Pham, and Christopher~D Manning. 2015.
\newblock Effective approaches to attention-based neural machine translation.
\newblock \emph{arXiv preprint arXiv:1508.04025}.

\bibitem[{Mallinson et~al.(2017)Mallinson, Sennrich, and
  Lapata}]{mallinson2017paraphrasing}
Jonathan Mallinson, Rico Sennrich, and Mirella Lapata. 2017.
\newblock Paraphrasing revisited with neural machine translation.
\newblock In \emph{Proceedings of the 15th Conference of the European Chapter
  of the Association for Computational Linguistics: Volume 1, Long Papers},
  pages 881--893.

\bibitem[{Mayhew et~al.(2020)Mayhew, Bicknell, Brust, McDowell, Monroe, and
  Settles}]{staple20}
Stephen Mayhew, Klinton Bicknell, Chris Brust, Bill McDowell, Will Monroe, and
  Burr Settles. 2020.
\newblock Simultaneous translation and paraphrase for language education.
\newblock In \emph{Proceedings of the ACL Workshop on Neural Generation and
  Translation (WNGT)}. ACL.

\bibitem[{Ott et~al.(2019)Ott, Edunov, Baevski, Fan, Gross, Ng, Grangier, and
  Auli}]{ott2019Fairseq}
Myle Ott, Sergey Edunov, Alexei Baevski, Angela Fan, Sam Gross, Nathan Ng,
  David Grangier, and Michael Auli. 2019.
\newblock fairseq: A fast, extensible toolkit for sequence modeling.
\newblock \emph{arXiv preprint arXiv:1904.01038}.

\bibitem[{Ott et~al.(2018)Ott, Edunov, Grangier, and Auli}]{ott2018scaling}
Myle Ott, Sergey Edunov, David Grangier, and Michael Auli. 2018.
\newblock Scaling neural machine translation.
\newblock \emph{arXiv preprint arXiv:1806.00187}.

\bibitem[{Pang et~al.(2003)Pang, Knight, and Marcu}]{pang2003syntax}
Bo~Pang, Kevin Knight, and Daniel Marcu. 2003.
\newblock Syntax-based alignment of multiple translations: Extracting
  paraphrases and generating new sentences.
\newblock In \emph{Proceedings of the 2003 Conference of the North American
  Chapter of the Association for Computational Linguistics on Human Language
  Technology-Volume 1}, pages 102--109. Association for Computational
  Linguistics.

\bibitem[{Poliak et~al.(2018)Poliak, Belinkov, Glass, and
  Van~Durme}]{poliak2018evaluation}
Adam Poliak, Yonatan Belinkov, James Glass, and Benjamin Van~Durme. 2018.
\newblock On the evaluation of semantic phenomena in neural machine translation
  using natural language inference.
\newblock \emph{arXiv preprint arXiv:1804.09779}.

\bibitem[{Prakash et~al.(2016)Prakash, Hasan, Lee, Datla, Qadir, Liu, and
  Farri}]{prakash2016neural}
Aaditya Prakash, Sadid~A Hasan, Kathy Lee, Vivek Datla, Ashequl Qadir, Joey
  Liu, and Oladimeji Farri. 2016.
\newblock Neural paraphrase generation with stacked residual lstm networks.
\newblock \emph{arXiv preprint arXiv:1610.03098}.

\bibitem[{Rozis and Skadin{\v{s}}(2017)}]{rozis2017tilde}
Roberts Rozis and Raivis Skadin{\v{s}}. 2017.
\newblock Tilde model-multilingual open data for eu languages.
\newblock In \emph{Proceedings of the 21st Nordic Conference on Computational
  Linguistics, NoDaLiDa, 22-24 May 2017, Gothenburg, Sweden}, 131, pages
  263--265. Link{\"o}ping University Electronic Press.

\bibitem[{Sennrich et~al.(2015)Sennrich, Haddow, and
  Birch}]{sennrich2015neural}
Rico Sennrich, Barry Haddow, and Alexandra Birch. 2015.
\newblock Neural machine translation of rare words with subword units.
\newblock \emph{arXiv preprint arXiv:1508.07909}.

\bibitem[{Skadi{\c{n}}{\v{s}} et~al.(2014)Skadi{\c{n}}{\v{s}}, Tiedemann,
  Rozis, and Deksne}]{skadicnvs2014billions}
Raivis Skadi{\c{n}}{\v{s}}, J{\"o}rg Tiedemann, Roberts Rozis, and Daiga
  Deksne. 2014.
\newblock Billions of parallel words for free: Building and using the eu
  bookshop corpus.
\newblock In \emph{Proceedings of LREC}.

\bibitem[{Soares et~al.(2018)Soares, Moreira, and Becker}]{soares2018large}
Felipe Soares, Viviane Moreira, and Karin Becker. 2018.
\newblock A large parallel corpus of full-text scientific articles.
\newblock In \emph{Proceedings of the Eleventh International Conference on
  Language Resources and Evaluation (LREC-2018)}.

\bibitem[{Steinberger et~al.(2013)Steinberger, Eisele, Klocek, Pilos, and
  Schl{\"u}ter}]{steinberger2013dgt}
Ralf Steinberger, Andreas Eisele, Szymon Klocek, Spyridon Pilos, and Patrick
  Schl{\"u}ter. 2013.
\newblock Dgt-tm: A freely available translation memory in 22 languages.
\newblock \emph{arXiv preprint arXiv:1309.5226}.

\bibitem[{Steinberger et~al.(2006)Steinberger, Pouliquen, Widiger, Ignat,
  Erjavec, Tufis, and Varga}]{steinberger2006jrc}
Ralf Steinberger, Bruno Pouliquen, Anna Widiger, Camelia Ignat, Tomaz Erjavec,
  Dan Tufis, and D{\'a}niel Varga. 2006.
\newblock The jrc-acquis: A multilingual aligned parallel corpus with 20+
  languages.
\newblock \emph{arXiv preprint cs/0609058}.

\bibitem[{Sutskever et~al.(2014)Sutskever, Vinyals, and
  Le}]{sutskever2014sequence}
Ilya Sutskever, Oriol Vinyals, and Quoc~V Le. 2014.
\newblock Sequence to sequence learning with neural networks.
\newblock In \emph{Advances in neural information processing systems}, pages
  3104--3112.

\bibitem[{Tiedemann(2012)}]{OPUS}
J{\"o}rg Tiedemann. 2012.
\newblock Parallel data, tools and interfaces in opus.
\newblock 2012:2214--2218.

\bibitem[{Wieting and Gimpel(2017)}]{wieting2017paranmt}
John Wieting and Kevin Gimpel. 2017.
\newblock Paranmt-50m: Pushing the limits of paraphrastic sentence embeddings
  with millions of machine translations.
\newblock \emph{arXiv preprint arXiv:1711.05732}.

\bibitem[{Wo{\l}k and Marasek(2014)}]{wolk2014building}
Krzysztof Wo{\l}k and Krzysztof Marasek. 2014.
\newblock Building subject-aligned comparable corpora and mining it for truly
  parallel sentence pairs.
\newblock \emph{Procedia Technology}, 18:126--132.

\bibitem[{Yu et~al.(2018)Yu, Dohan, Luong, Zhao, Chen, Norouzi, and
  Le}]{yu2018qanet}
Adams~Wei Yu, David Dohan, Minh-Thang Luong, Rui Zhao, Kai Chen, Mohammad
  Norouzi, and Quoc~V Le. 2018.
\newblock Qanet: Combining local convolution with global self-attention for
  reading comprehension.
\newblock \emph{arXiv preprint arXiv:1804.09541}.

\bibitem[{Zarrabi-Zadeh(2007)}]{zarrabizadeh2007tanzil}
Hamid Zarrabi-Zadeh. 2007.
\newblock Tanzil project.
\newblock \emph{URL: http://tanzil. net/wiki/Tanzil\_Project}.

\end{thebibliography}
\bibliographystyle{acl_natbib}

\end{document}